%% file: Formatting-Instructions-LaTeX-2026.tex
\documentclass[letterpaper]{article} 
\usepackage{aaai2026}  
\usepackage{times}  
\usepackage{helvet}  
\usepackage{courier}  
\usepackage[hyphens]{url}  
\usepackage{graphicx} 
\usepackage{natbib}  
\usepackage{caption} 
\usepackage{amsmath}       
\usepackage{siunitx}      
\usepackage[ruled,vlined]{algorithm2e}  
\usepackage{amssymb}
\usepackage{booktabs}

\usepackage{multirow}
\usepackage[table,xcdraw]{xcolor}  
\usepackage{tabularx}  
\usepackage{pifont}
\usepackage{array}

\newcommand{\appen}[1]{Appendix {#1}}

\def\ourmethod{Anomagic}
\def\ourdataset{AnomVerse}

\title{Anomagic: Crossmodal Prompt-driven Zero-shot Anomaly Generation}
\author{    
Yuxin Jiang\textsuperscript{\rm 1},    
Wei Luo\textsuperscript{\rm 3},    
Hui Zhang\textsuperscript{\rm 2},    
Qiyu Chen\textsuperscript{\rm 4},    
Haiming Yao\textsuperscript{\rm 3},\\    
Weiming Shen\thanks{corresponding author.}\textsuperscript{\rm 1},    
Yunkang Cao$^*$ \textsuperscript{\rm 2}
}
\affiliations{    
\textsuperscript{\rm 1}Huazhong University of Science and Technology,    
\textsuperscript{\rm 2}Hunan University,    
\textsuperscript{\rm 3}Tsinghua University\\    
\textsuperscript{\rm 3}Institute of Automation, Chinese Academy of Sciences\\
yuxinjiang@hust.edu.cn, luow23@mails.tsinghua.edu.cn, zhanghuihby@126.com, chenqiyu2021@ia.ac.cn, yhm22@mails.tsinghua.edu.cn, wshen@ieee.org, caoyunkang@ieee.org, 
}

\usepackage{subcaption}
\usepackage{amsfonts}

\begin{document}

\maketitle

\begin{abstract}
We propose \ourmethod{}, a zero-shot anomaly generation method that produces semantically coherent anomalies without requiring any exemplar anomalies. By unifying both visual and textual cues through a crossmodal prompt encoding scheme, \ourmethod{} leverages rich contextual information to steer an inpainting‐based generation pipeline. A subsequent contrastive refinement strategy enforces precise alignment between synthesized anomalies and their masks, thereby bolstering downstream anomaly detection accuracy. To facilitate training, we introduce \ourdataset{}, a collection of 12,987 anomaly–mask–caption triplets assembled from 13 publicly available datasets, where captions are automatically generated by multimodal large language models using structured visual prompts and template‐based textual hints. Extensive experiments demonstrate that \ourmethod{} trained on \ourdataset{} can synthesize more realistic and varied anomalies than prior methods, yielding superior improvements in downstream anomaly detection. Furthermore, \ourmethod{} can generate anomalies for any normal‐category image using user‐defined prompts, establishing a versatile foundation model for anomaly generation.
\end{abstract}

\begin{links}
    \link{Code}{https://github.com/yuxin-jiang/Anomagic}
\end{links}

\input{source/1-introduction}

\input{source/2-related-work}

\input{source/3-method}

\input{source/4-experiment}
\input{source/5-conclusion}
\input{source/Acknowledgments}


\input{aaai2026.bbl}
\end{document}

%% file: source/1-introduction.tex
\section{Introduction}

Anomalies, characterized as rare or absent patterns in typical conditions~\cite{xgy_survey,cao_survey}, pose significant challenges in domains such as manufacturing, where defects can precipitate critical safety concerns~\cite{patchcore}. To address this, unsupervised anomaly detection has emerged as a pivotal approach, training models exclusively on normal data to identify deviations~\cite{RD,INP-Former}. Concurrently, efforts to mitigate the scarcity of anomaly samples have led to advancements in anomaly generation, aiming to produce realistic and diverse anomalous instances~\cite{DefectFill,DualAnoDiff}.

However, despite their potential, the majority of current anomaly generation methods, including DualAnoDiff~\cite{DualAnoDiff} and AnoGen~\cite{AnoGen}, function within a \emph{few-shot anomaly generation} paradigm. This approach depends on a limited set of annotated anomalies from known defect categories. Although these methods can effectively generate visually plausible anomalies within familiar classes, they are unable to generalize to unseen defect types or new object categories. The task of generating realistic anomalies for novel categories without prior examples, known as \emph{zero-shot anomaly generation}~\cite{AnomalyAny}, remains a significant and largely unaddressed challenge.

Initial efforts in zero-shot anomaly generation adopted a cut-and-paste methodology~\cite{Draem,realnet}, overlaying external textures onto normal images. Although these approaches yielded diverse anomalies, their synthetic appearance often undermined practical utility. A notable advancement was achieved with AnomalyAny~\cite{AnomalyAny}, which leverages pre-trained Stable Diffusion (SD) models~\cite{SD} guided by textual prompts. By manipulating the attention matrix during the denoising process, AnomalyAny achieves versatility across diverse image categories. However, its attention mechanism~\cite{chefer2023attend}, limited to a small set of tokens, struggles to capture the semantic richness required for highly diverse anomalies. Furthermore, insights from other image generation domains~\cite{dreambooth,dang2025personalized} suggest that task-specific fine-tuning can significantly enhance the quality of generated images.

In this work, we present \textbf{\ourmethod{}}, an innovative framework for zero-shot anomaly generation that incorporates crossmodal prompts, merging visual and textual semantics to create anomalies that are both highly realistic and adaptable. This method enables guidance through visual inputs, textual inputs, or a combination of both, thereby ensuring versatility across a wide range of scenarios. Furthermore, we introduce a contrastive anomaly mask refinement strategy, which produces precise mask-anomaly pairs, significantly improving upon the coarse masks generated by previous methods. For the training of \ourmethod{}, we develop a crossmodal prompting strategy that utilizes multimodal large language models (MLLMs) to generate accurate captions for anomaly images, informed by both visual and textual data. These captions ensure that the generated anomalies closely match their intended semantic descriptions. By applying this strategy to publicly available datasets such as MVTec AD~\cite{MVTec-AD} and VisA~\cite{VisA}, we have created \textbf{\ourdataset{}}, which, with 12,987 anomaly-mask-caption triplets, stands as the largest dataset of its kind to date.

When trained on \ourdataset{} using a novel prompt-guided inpainting approach, \ourmethod{} achieves remarkable levels of realism and diversity in the anomalies it generates, while also improving the performance of subsequent anomaly detection tasks. Our experimental findings show that \ourmethod{} not only generates anomalies that accurately reflect the prompts from \ourdataset{} but also demonstrates strong generalization capabilities, allowing it to handle arbitrary prompts and generate corresponding anomalies for entirely new categories. This positions \ourmethod{} as a foundational model for zero-shot anomaly generation. The key contributions of this work are as follows:
\begin{itemize}
    \item We introduce \ourmethod{}, a novel framework for zero-shot anomaly generation that utilizes crossmodal prompts to achieve superior realism, supported by a contrastive mask refinement strategy that ensures the precision of anomaly masks.
    \item We present \ourdataset{}, an extensive dataset consisting of 12,987 anomaly-mask-caption triplets, developed through a novel crossmodal prompting technique.
    \item Rigorous experimental validation illustrating that \ourmethod{} outperforms existing methods in generating high-quality anomalies and bolstering anomaly detection performance, while demonstrating strong generalization across diverse prompts, including unimodal, crossmodal, and user-defined inputs.
\end{itemize}

%% file: source/2-related-work.tex
\section{Related Works}

\subsection{Anomaly Generation}

\noindent\textbf{Few-shot Anomaly Generation.}
Few-shot anomaly generation methods leverage a small set of abnormal exemplars during training to synthesize novel anomalies of comparable types, thereby enriching the diversity of anomaly categories. Early work such as Defect‑GAN~\cite{Defect-GAN} and DFMGAN~\cite{DFMGAN} adopted Generative Adversarial Networks (GANs)~\cite{GAN} for their demonstrated ability to produce high‑fidelity images. More recent approaches have transitioned to diffusion‑based frameworks: AnomalyDiffusion~\cite{anomalydiffusion} employs text inversion~\cite{TexTInversion} to capture anomaly semantics and mask distributions, while Defect‑Gen~\cite{Defect_Spectrum} introduces a two‑stage diffusion process to improve realism. Methods such as AnoGen~\cite{AnoGen} and DefectFill~\cite{DefectFill} formulate anomaly synthesis as an inpainting task to guarantee mask consistency, and DualAnoDiff~\cite{DualAnoDiff} further refines generation quality by alternately modeling foreground and background components. SeaS~\cite{SeaS} extends this paradigm by binding anomaly attributes to distinct prompt tokens, enabling a single model to generate multiple anomaly types. Despite these advances, all of these techniques remain limited to the anomaly types seen during training, restricting their capacity to generalize to unseen defect classes.

\noindent\textbf{Zero-shot Anomaly Generation.}
Zero-shot anomaly generation aims to produce realistic anomalies for categories not encountered during training, without the need for prior abnormal samples. Initial approaches, such as CutPaste~\cite{Cutpaste} and DRAEM~\cite{Draem}, depended on rudimentary cut-and-paste techniques, superimposing external textures onto normal images. While these methods provided diversity, the generated anomalies frequently lacked authenticity. A notable advancement was made by AnomalyAny~\cite{AnomalyAny}, which leveraged pre-trained SD models~\cite{SD} directed by textual prompts, modifying the attention matrix during the denoising process to accommodate a variety of image categories. Nonetheless, its reliance on single-modal prompts compromise its controllability. Our proposed method, {\ourmethod{}}, addresses these limitations by incorporating crossmodal prompts to enrich semantic understanding and by expediting the generation process.

\subsection{Conditional Diffusion Models}

\noindent\textbf{Crossmodal Conditions.}
Diffusion models conditioned on multiple modalities have substantially advanced the controllability of generative processes by aligning outputs with diverse guidance, such as textual prompts~\cite{dhariwal2021diffusion,SD} and visual cues~\cite{ControlNet}. More recent frameworks—e.g., BLIP-Diffusion~\cite{Blip-diffusion} and OmniGen~\cite{xiao2025omnigen}—jointly leverage language and visual embeddings to achieve finer-grained semantic alignment. Unlike earlier anomaly synthesis approaches that rely solely on text-driven control, \ourmethod{} exploits a crossmodal conditioning scheme that seamlessly integrates visual representations of defects with rich textual descriptions to produce highly targeted anomalous examples.

\noindent\textbf{Mask Generation.}
Accurate pixel‐level masks are indispensable for anomaly detection~\cite{AnoGen,AnomalyAny}. Traditional mask‐aware synthesis techniques often depend on heuristic post‐processing to extract imperfect region proposals~\cite{nguyen2023dataset,qian2024maskfactory}, which falls short of the pixel-level fidelity required in practice. Approaches such as DefectFill~\cite{DefectFill} and AnomalyDiffusion~\cite{anomalydiffusion} constrain generated anomalies to rough mask shapes, while AnomalyAny derives masks from attention maps between text tokens and latent features—yet this tends to yield coarse boundaries. In contrast, \ourmethod{} introduces a contrastive mask refinement module that accuraly derive anomaly masks via discrepancies between input normal image and generated anomalies, yielding pixel‐accurate masks suitable for downstream anomaly detection.

%% file: source/3-method.tex
\section{Dataset: \ourdataset{}}
\subsection{Construction Pipeline}
We introduce a novel pipeline, as illustrated in Figure~\ref{fig:dataset}(a), designed to curate high-quality triplets comprising anomalies, masks, and captions to facilitate the training of zero-shot anomaly generation models. Although anomalies and their associated masks are widely available in public datasets, generating precise and informative captions for anomaly regions poses a significant challenge. To address this, we employ a crossmodal prompting strategy that leverages MLLMs, specifically the Doubao-Seed-1.6-thinking model\footnote{\url{https://www.volcengine.com/product/doubao}}, to produce detailed captions for the anomalies. 
To improve the quality and relevance of these captions, we utilize masks to delineate minimal bounding boxes to highlight defective areas. Additionally, we have devised a structured caption template: ``The image depicts [general description of the object], with a [type of defect] observed [location description]. The defect is characterized by [detailed description] and exhibits [notable features].'' This template ensures consistency and clarity in the caption generation process. By combining visual prompts, in the form of bounding boxes, with this predefined textual template, our method enables the creation of precise anomaly-mask-caption triplets.

\subsection{Statistics}
The \ourdataset{} dataset integrates data from 13 publicly accessible datasets, including MVTec AD \cite{MVTec-AD}, VisA \cite{VisA}, and MANTA~\cite{MANTA}, et al,  spanning five distinct domains (Figure~\ref{fig:dataset}(b)): industrial (56.5\%), textiles (23.6\%), consumer goods (8.7\%), medicine (5.9\%), and electronics (5.3\%). Comprising 12,987 anomaly samples that represent 131 defect types, each sample is paired with a structured descriptive caption. In contrast to its predecessor, MMAD~\cite{MMAD}, which includes 8,366 samples, \ourdataset{} provides a significantly larger collection, establishing it as a critical resource for advancing research in zero-shot anomaly generation. Further details and examples of \ourdataset{} are provided in \appen{A}.

\section{Method: \ourmethod{}}
\subsection{Preliminaries}

\noindent\textbf{Latent Diffusion Models.}  
Latent Diffusion Models (LDMs)~\citep{rombach2022high} constitute a class of diffusion-based generative models~\citep{ho2020denoising,dhariwal2021diffusion,song2020score} that operate within a reduced-dimensional latent space, thereby mitigating computational demands. Initially, an encoder $E$ transforms an image $\mathbf{I}$ into a latent representation $\mathbf{z}_0 = E(\mathbf{I})$. During the generative process, this latent representation undergoes a diffusion process involving noise addition and denoising before being reconstructed by a decoder $D$. Specifically, during the training phase, Gaussian noise $\varepsilon \sim \mathcal{N}(0, \mathbb{I})$ is introduced to $\mathbf{z}_0$ according to a predefined schedule $\{\alpha_t\}_{t=1}^T$, resulting in the noisy latent code $\mathbf{z}_t = \sqrt{\alpha_t} \mathbf{z}_0 + \sqrt{1 - \alpha_t} \varepsilon$. A neural network $\varepsilon_\theta$ is then trained to predict the noise component $\varepsilon$ from the noisy latent code $\mathbf{z}_t$ and the timestep $t$. Through the reverse diffusion process, this noise prediction enables the recovery of a reconstructed latent representation $\hat{\mathbf{z}}_0$. Finally, the image is reconstructed by the decoder $D$ as $\hat{\mathbf{I}} = D(\hat{\mathbf{z}}_0)$. The corresponding loss function is given by: 
\begin{equation}
\mathcal{L}_{\mathrm{LDM}} = \mathbb{E}_{\mathbf{z}_0, \varepsilon, t} \left\| \varepsilon - \varepsilon_\theta(\mathbf{z}_t, t) \right\|_2^2.
\end{equation}

\noindent\textbf{Conditional Diffusion Models.}  
To facilitate the generation of images with desired semantic attributes, conditional diffusion models extend LDMs by incorporating supplementary inputs—such as textual descriptions or masks—through cross-attention mechanisms~\citep{SD,ControlNet}. Specifically, a condition embedding $\mathbf{P}$ is integrated into the UNet's latent feature maps at various scales, thereby aligning the denoising process with the provided guidance. This ensures that the synthesized images conform to the specified conditions. The training objective is accordingly adjusted as follows:
\begin{equation}
\mathcal{L}_{\mathrm{LDM}} = \mathbb{E}_{\mathbf{z}_0, \varepsilon, t} \left\| \varepsilon - \varepsilon_\theta(\mathbf{z}_t, t, \mathbf{P}) \right\|_2^2.
\end{equation}

\input{figures/fig_dataset}

\input{figures/fig_method}

\subsection{Framework Overview}
Despite the advancements in conditional LDMs, generating anomalies in industrial contexts presents two primary challenges: (1) formulating precise, defect-specific conditions, and (2) ensuring generated anomalies align with real-world industrial characteristics while maintaining contextual coherence. To address these issues, our proposed framework, \ourmethod{}, as depicted in Figure~\ref{fig:method}, employs Crossmodal Prompt Encoding (CPE) to derive fine-grained conditions and fine-tunes pre-trained SD models using Low-Rank Adaptation (LoRA)~\citep{LoRA} in conjunction with an inpainting strategy. During the testing phase, we utilize MLLMs to semantically retrieve the most relevant prompts from external sources, such as \ourdataset{}, and subsequently apply these conditions to generate anomalies on novel normal images in a zero-shot manner.

\subsection{Crossmodal Prompt Encoding}
Both visual and textual modalities possess rich semantic information that can reduce ambiguity while maintaining flexibility. To exploit these characteristics, we introduce the CPE scheme that extracts semantics from crossmodal prompts, including an anomalous image $\mathbf{I}^{\text{ref}}$ as a visual prompt, an anomaly mask $\mathbf{M}^{\text{ref}}$, and corresponding anomaly captions $\mathbf{t}^{\text{ref}}$ as textual prompts.

\noindent\textbf{Region-focused Visual Guidance.}  
To effectively capture visual cues pertinent to anomalies, we leverage a pre-trained CLIP~\citep{CLIP} image encoder to process the reference image $\mathbf{I}^{\text{ref}}$, yielding a feature map $\mathbf{F}$. However, directly applying the encoder to the entire image may overlook subtle defects due to the dominance of normal object semantics. To counteract this, we introduce a region-focused attention mechanism that isolates anomaly-specific representations by applying the binary anomaly mask $\mathbf{M}^{\text{ref}}$,
\begin{equation}\label{eq:visual_prompt}
\mathbf{P}_v = \text{Softmax}\left( \frac{\mathbf{Q} \mathbf{K}^T}{\sqrt{D}} - (1 - \mathbf{M}^{\text{ref}}) \cdot C \right) \mathbf{V},
\end{equation}
where $\mathbf{Q} = \theta_q(\mathbf{F})$, $\mathbf{K} = \theta_k(\mathbf{F})$, and $\mathbf{V} = \theta_v(\mathbf{F})$ are the query, key, and value projections, respectively. The constant $C$ serves as a large scaling factor to attenuate attention weights in masked regions. This approach diminishes the influence of background contexts by assigning negligible attention scores to masked regions. 

\noindent\textbf{Fine-grained Textual Semantics.}  
To accommodate verbose technical captions that exceed standard token limits (e.g., CLIP's 77-token constraint~\cite{CLIP}), we implement a hierarchical encoding strategy for long texts. The input text $\mathbf{t}^{\text{ref}}$ is divided into semantically coherent segments, each of which is encoded by CLIP. These encodings are then aggregated through mean-pooling to form a comprehensive global embedding $\mathbf{P}_t$. This method preserves long-range dependencies and enables precise control over textual prompts.

\noindent\textbf{Crossmodal Prompt Fusion.}  
We further develop a crossmodal fusion architecture to process visual features $\mathbf{P}_v$ and textual embeddings $\mathbf{P}_t$ through modality-specific cross-attention blocks, capturing bidirectional interactions between the modalities. The resulting unified crossmodal semantic feature $\mathbf{P}_c$ serves as the conditioning input for the diffusion process:
\begin{equation}\label{eq:crossmodal_prompt}
\mathbf{P}_c = \text{CrossFusion}(\mathbf{P}_v,  \mathbf{P}_t),
\end{equation}
where $\text{CrossFusion}(\cdot)$ represents a composite function integrating projection networks for $\mathbf{P}_v$ and $\mathbf{P}_t$ alignment, cross-attention, and fusion operations. Only the attention module within visual guidance and the $\text{CrossFusion}(\cdot)$ module are optimized during training, with the pre-trained CLIP model kept frozen to preserve its generalization capacity. The trainable parameters are collectively denoted as $\theta_{CPE}$.

\input{tables/algorithm_training}

\subsection{Training via Prompt-guided Inpainting}
To achieve precise anomaly generation that adheres to both semantic and spatial constraints, we fine-tune a pre-trained inpainting SD model using LoRA on its cross-attention layers. This efficient fine-tuning approach enables the model to generate defects that align with the crossmodal prompts while respecting the specified spatial regions. In each training iteration, we sample a triplet consisting of $\mathbf{I}^{\text{ref}}$,  $\mathbf{M}^{\text{ref}}$, and  $\mathbf{t}^{\text{ref}}$ from \ourdataset{}.
Subsequently, $\mathbf{M}^{\text{ref}}$ undergoes dilation to produce an expanded inpainting mask, $\mathbf{M}_{\text{inpainting}}$, within which the pixels of the  $\mathbf{I}^{\text{ref}}$ are masked to produce the input image $\mathbf{I}_{\text{input}}$ for the denoising process.
During denoising, we compel the diffusion model to inpaint the masked regions with the original anomalies by minimizing the noise prediction error, while ensuring that the non-masked regions follow the original image's denoising trajectory. This is formalized in the modified loss function:
\begin{equation}\label{eq:masked_loss}
\mathcal{L}_{\text{LDM}}' = \mathbb{E}_{\mathbf{z}_0, \varepsilon, t} \left\| \left( \varepsilon - \varepsilon_\theta(\mathbf{z}_t, t, \mathbf{P}_c) \right) \odot \mathbf{M}_{\text{inpainting}} \right\|_2^2.
\end{equation}
The training protocol is detailed in Algorithm~\ref{alg:training}.

\subsection{Inference}

\noindent\textbf{Prompt-driven Anomaly Generation.}  
Our framework facilitates zero-shot anomaly synthesis by leveraging arbitrary anomaly-mask-caption triplets $(\mathbf{I}^{\text{ref}}, \mathbf{M}^{\text{ref}}, \mathbf{t}^{\text{ref}})$. During the generation phase, we initially extract crossmodal features from the given triplet using the CPE scheme, as detailed in Equation~\eqref{eq:crossmodal_prompt}, resulting in the crossmodal condition $\mathbf{P}_c$. Subsequently, a coarse anomaly mask, $\mathbf{M}_{\text{inpainting}}$, is randomly sampled. The diffusion model then synthesizes anomalies on the target image $\mathbf{I}$ by updating the latent variables exclusively within the masked regions, thereby preserving the integrity of normal areas. Although our method is capable of accommodating arbitrary prompts, we have implemented an automated pipeline by default to streamline the process. This pipeline selects the most pertinent prompts from \ourdataset{}, allowing users to simply provide a user query $Q$, such as ``What defects commonly appear in cashews?'' Our system then employs an MLLM to generate a semantic response $A$, for instance, ``cracks, holes, bulges, scratches.'' Then we retrieve semantically aligned anomaly categories from \ourdataset{} using the MLLM, and select the corresponding anomaly-mask-caption triplets. Additional results regarding the retrieval are elaborated in \appen{B}.

\noindent\textbf{Contrastive Anomaly Mask Refinement.}  
Considering that the synthesized anomalies might not entirely fill the initial coarse mask, we introduce a contrastive anomaly mask refinement strategy for better alignment. As illustrated in Figure~\ref{fig:method}, our inpainting-based generation approach ensures that discrepancies occur only in anomaly regions. Thus, we compute pixel-level differences between input and output images and apply a threshold to obtain refined binary masks, denoted as $\mathbf{M}_{\text{r}}$. To achieve this, we utilize a pre-trained MetaUAS~\citep{MetaUAS} model, which is specifically designed to detect pixel-level discrepancies between two images, with a threshold of 0.9. As shown in Figure~\ref{fig:qualitative_results}, our refinement step provides better alignment between the anomalies and masks.

%% file: figures/fig_dataset.tex
\begin{figure}[t!]
\centering
\includegraphics[width=0.9\linewidth]{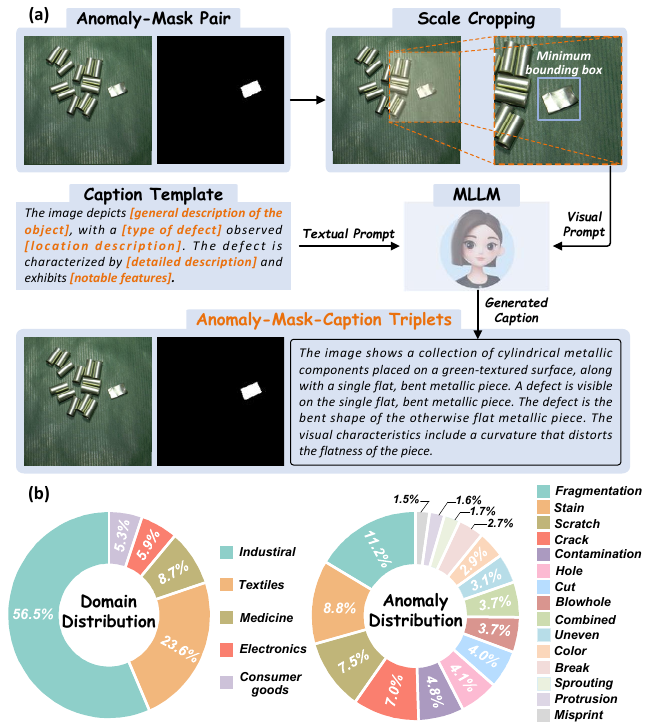}
\caption{\textbf{Overview of \ourdataset{}.} (a) The construction pipeline for generating anomaly–mask–caption triplets. (b) The distribution across domains and the top 15 most frequent anomaly categories.}
\label{fig:dataset}
\end{figure}

%% file: figures/fig_method.tex
\begin{figure*}[h!]
\centering\includegraphics[width=.9\linewidth]
{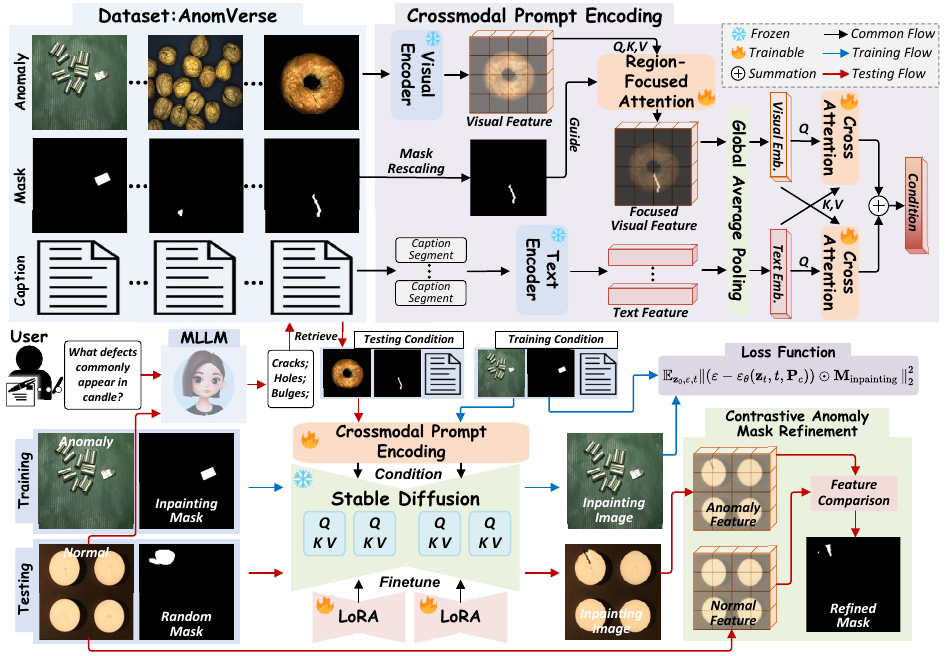}
\caption{\textbf{Overall framework of \ourmethod{}.} Our method employs a crossmodal prompt encoding to extract conditions from selected anomaly–mask–caption triplets in \ourdataset{}, which are then used to guide the inpainting process. During testing, a contrastive anomaly mask refinement module is introduced to further enhance the accuracy of the predicted anomaly masks.}
\label{fig:method}
\end{figure*} 

%% file: tables/algorithm_training.tex
\begin{algorithm}[t]
\caption{Training process of \ourmethod{}}
\label{alg:training}
\SetAlgoLined
\KwIn{Reference triplets $(\mathbf{I}^{\mathrm{ref}}, \mathbf{M}^{\mathrm{ref}}, \mathbf{t}^{\mathrm{ref}})$}
\KwOut{LoRA weights $\theta_L$ and trainable CPE weights $\theta_{\mathrm{CPE}}$}
\BlankLine
\textbf{1. Initialization:}\\
Load pretrained SD and CLIP parameters.\\
Initialize $\theta_L,\theta_{\mathrm{CPE}}$.\\
\textbf{2. Main Loop:}\\
\ForEach{sample $(\mathbf{I}^{\mathrm{ref}}, \mathbf{M}^{\mathrm{ref}}, \mathbf{t}^{\mathrm{ref}})$}{
  \emph{(a) Crossmodal Prompt Encoding:}\\
  Compute prompt features $\mathbf{P}_c$ via Equations~\eqref{eq:visual_prompt}–\eqref{eq:crossmodal_prompt}.\\
  \emph{(b) Input Preparation:}\\
  Dilate  $\mathbf{M}^{\mathrm{ref}}$ to get the inpainting mask  $\mathbf{M}_{\text{inpainting}}$. \\
  Mask $\mathbf{I}^{\mathrm{ref}}$ with  $\mathbf{M}_{\text{inpainting}}$ to get  $\mathbf{I}_{\text{input}}$.\\
    \emph{(c) Compute Loss and Update:}\\
  \For{$t\sim\mathcal{U}[1,T]$}{
   Add noise into the embedding of $\mathbf{I}_{\text{input}}$ to obtain $\mathbf{z}_t$.\\
   Predict noise with $\varepsilon_\theta(\mathbf{z}_t,t,\mathbf{P}_c)$.\\
  Compute masked denoising loss $\mathcal{L}_{\text{LDM}}'$ via Eq.~\eqref{eq:masked_loss}.\\
  Backpropagate $\mathcal{L}_{\text{LDM}}'$ to update $\theta_L,\theta_{\mathrm{CPE}}$.
  } 
}
\Repeat{convergence}{continue}
\end{algorithm}

%% file: source/4-experiment.tex
\input{figures/fig_qualitative_results}

\section{Experiments}

\subsection{Experimental Settings}
\noindent\textbf{Dataset.} Our \ourdataset{} dataset integrates data from 13 publicly accessible datasets. For evaluation, we predominantly utilize the MVTec AD and VisA datasets, while the remaining 11 datasets are reserved for training purposes. The VisA dataset~\cite{VisA} constitutes the primary resource for our experimental investigations, with supplementary results from MVTec AD~\cite{MVTec-AD} detailed in \appen{E}.

\noindent\textbf{Implementation details.}
Our proposed \ourmethod{} is implemented using Diffusers~\cite{diffusers2022}, based on Stable Diffusion v1.5 with OpenCLIP ViT-H/14~\cite{openclip2021}. We adopt a 20-step DDIM sampler~\cite{DDIM} to balance generation quality and computational efficiency. For training downstream anomaly detection methods, we generate one anomalous image per normal image.

\noindent\textbf{Metrics.} For anomaly generation, we adopt the Inception Score (IS) and the Intra-cluster LPIPS distance (IL)~\cite{anomalydiffusion}. For anomaly detection and localization, we report the image-level area under the ROC curve (I-ROC) and maximum F1 score (I-F1), as well as the pixel-level area under the per-region overlap curve (PRO) and maximum F1 score (P-F1)~\cite{INP-Former}.

\subsection{Comparison in Anomaly Generation}

\noindent\textbf{Baselines.} We primarily compare our method against existing zero-shot anomaly generation techniques, namely DRAEM~\cite{Draem}, RealNet~\cite{realnet}, and AnomalyAny~\cite{AnomalyAny}. Additionally, we benchmark our approach against AnoGen~\cite{AnoGen}, a state-of-the-art (SOTA) few-shot anomaly generation method trained using one image per defect type per category from VisA.

\noindent\textbf{Quantitative Results.} As illustrated in Table~\ref{table:anomaly_generation_results}, our method surpasses all zero-shot anomaly generation methods in IS and IL, exceeding even AnoGen, which requires real anomalies for training. These results indicate that \ourmethod{} excels in both the realism and diversity of generated anomalies.

\input{tables/table1}

\noindent\textbf{Qualitative Comparisons.} 
Figure~\ref{fig:qualitative_results} demonstrates that our method produces anomalous samples with greater realism compared to cut-paste-based approaches, while also outperforming AnoGen and AnomalyAny in both generation quality and mask accuracy. Note that precise anomaly masks are critical for downstream anomaly detection training. Unlike AnomalyAny, which requires approximately 3 minutes to generate a single anomalous image, our end-to-end synthesis pipeline—including anomaly generation and mask refinement—achieves an average runtime of only 1.2 seconds per sample with a resolution of $512\times512$ on a single NVIDIA A100 Tensor Core GPU. \appen{D and E} further highlights the versatility of our approach across various domains, including medical imaging and web-crawled image data.

\noindent\textbf{Data Distribution Comparison.} 
Figure~\ref{fig:tsne} visualizes the feature distributions of normal samples, real anomalies, and synthesized anomalies using t-SNE with a pre-trained ResNet50 \cite{resnet}. As shown, the anomalies generated by \ourmethod{} exhibit a distribution closely aligned with that of real anomalies, outperforming the few-shot method AnoGen, suggesting that \ourmethod{} is capable of synthesizing anomalies that resemble real-world defect patterns.

\input{figures/fig_tsne}

\subsection{Comparison in Anomaly Detection}
\noindent\textbf{Evaluation Paradigm:} Unlike existing methods~\cite{anomalydiffusion,DualAnoDiff} that typically employ UNet for supervised training without comparing to SOTA anomaly detection techniques, our approach focuses on enhancing a recent SOTA method, INP-Former++~\cite{INP-Former++}, by integrating advanced anomaly generation techniques. More details about this paradigm are presented in \appen{C}.

\noindent\textbf{Quantitative Comparison.}
As shown in Table~\ref{tab:method_comparison}, our generated samples enhance anomaly detection performance in INP-Former++ on VisA, and even surpass AnoGen that requires real anomalies for training, with higher I-F1/PRO/P-F1 (96.77\%/95.92\%/54.00\%  vs 96.55\%/95.62\%/52.61\%).

\input{tables/table2}

\subsection{Ablation study}
\noindent\textbf{Overall Ablation.} As shown in Table~\ref{table:ablation_cpe_lora}, incorporating CPE and LoRA significantly improves both the quality of anomaly generation and the performance of downstream anomaly detection, yielding a 2.06\% increase in P-F1 score over the baseline DRAEM.

\input{tables/table3}

\noindent\textbf{Unimodal and Crossmodal Prompt Evaluation.}  
To assess the adaptability of \ourmethod{} in generating anomalies, we conducted experiments on VisA using both unimodal prompts—namely \ourmethod{}-Text and \ourmethod{}-Visual—and crossmodal prompts (\ourmethod{}-Cross). As shown in Figure~\ref{fig:crossmodal_prompts}, \ourmethod{} reliably produces high-fidelity anomalies even when provided with single-modality prompts. Moreover, anomalies synthesized by \ourmethod{}-Cross display enhanced realism, closely mirroring the texture, structure, and contextual coherence of genuine anomalous instances. These findings underscore the method’s flexibility, allowing users to employ either unimodal or crossmodal prompts according to their specific requirements.

\noindent\textbf{Anomaly Generation under User-Defined Prompts.}
Figure~\ref{fig:one_shot} illustrates our method's capability to generate realistic anomalies using arbitrary user-specified prompts, including those derived from real anomalies distinct from our training data. For example, although our model was not trained on VisA, it effectively produces convincing anomalies based on diverse prompts inspired by VisA. Furthermore, our method can also effectively generate realistic anomalies even with novel web-sourced prompts, as detailed in the \appen{E}. This positions it as a foundational framework for adaptable anomaly generation across diverse, user-defined prompts.

\input{figures/fig_crossmodal_prompt}
\input{figures/fig_one_shot}

%% file: figures/fig_qualitative_results.tex
\begin{figure*}[htbp]
\centering\includegraphics[width=.9\linewidth]{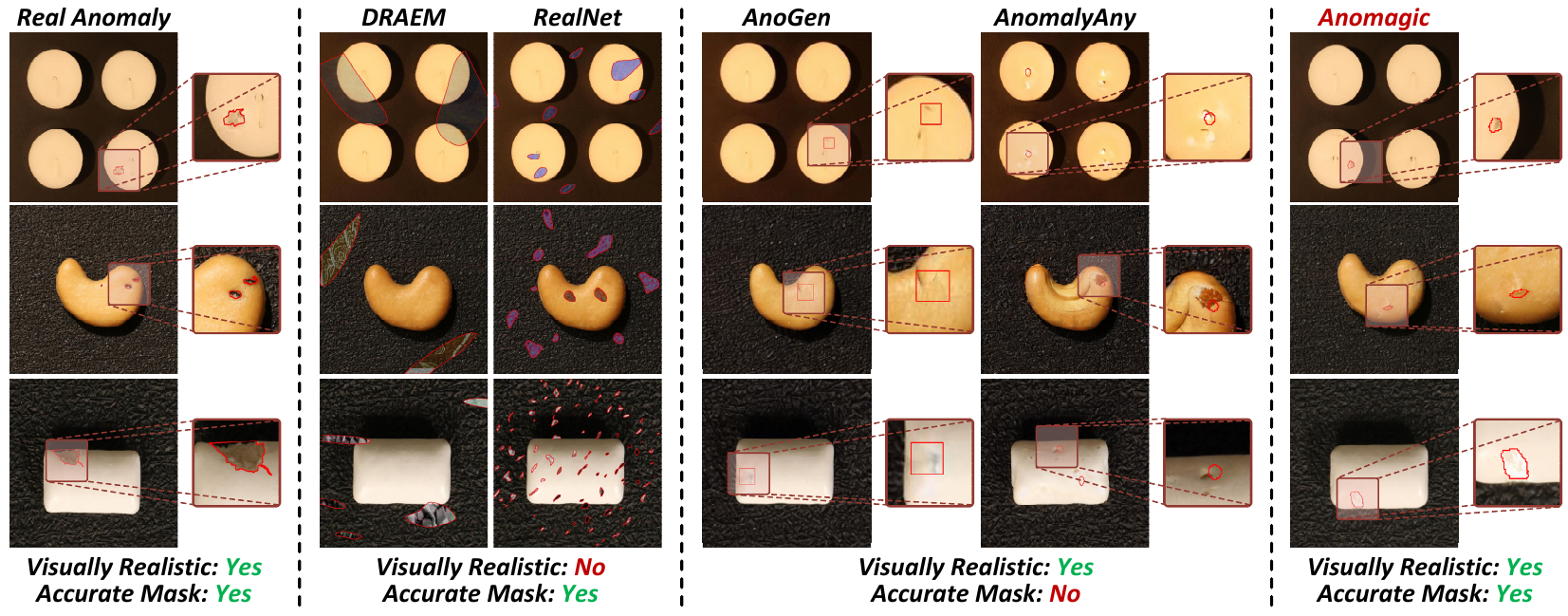}
\caption{\textbf{Qualitative comparison of anomaly generation performance.} Anomalies are highlighted with red circles. Unlike existing zero-shot methods (DRAEM and RealNet) and few-shot methods (AnoGen), \ourmethod{} uniquely achieves both visually realistic anomaly synthesis and accurate anomaly mask generation.}
\label{fig:qualitative_results}
\end{figure*}

%% file: tables/table1.tex
\begin{table}[!ht]
\centering
\setlength{\tabcolsep}{3pt}
\renewcommand{\arraystretch}{0.8} 
\fontsize{9}{10}\selectfont{
\begin{tabular}{c|clclclcl|
>{}c 
>{}l }
\toprule
Cate.    & \multicolumn{2}{c}{AnoGen}             & \multicolumn{2}{c}{DRAEM}     & \multicolumn{2}{c}{RealNet}   & \multicolumn{2}{c|}{AnoAny}         & \multicolumn{2}{c}{\textbf{Anomagic}}  \\ \midrule
pcb1        & \multicolumn{2}{c}{1.48/0.26}          & \multicolumn{2}{c}{1.35/0.24} & \multicolumn{2}{c}{1.38/0.23} & \multicolumn{2}{c|}{\textbf{1.64/0.10}} & \multicolumn{2}{c}{1.58/0.33}          \\
pcb2        & \multicolumn{2}{c}{\textbf{1.79/0.45}} & \multicolumn{2}{c}{1.33/0.43} & \multicolumn{2}{c}{1.32/0.40} & \multicolumn{2}{c|}{1.36/0.25}          & \multicolumn{2}{c}{1.70/0.42}          \\
pcb3        & \multicolumn{2}{c}{1.70/0.32}          & \multicolumn{2}{c}{1.53/0.28} & \multicolumn{2}{c}{1.46/0.27} & \multicolumn{2}{c|}{1.62/0.18}          & \multicolumn{2}{c}{\textbf{2.00/0.28}} \\
pcb4        & \multicolumn{2}{c}{1.46/0.42}          & \multicolumn{2}{c}{1.36/0.39} & \multicolumn{2}{c}{1.34/0.38} & \multicolumn{2}{c|}{1.46/0.38}          & \multicolumn{2}{c}{\textbf{1.62/0.41}} \\
maca.1   & \multicolumn{2}{c}{\textbf{2.64/0.41}} & \multicolumn{2}{c}{1.87/0.37} & \multicolumn{2}{c}{1.97/0.37} & \multicolumn{2}{c|}{1.81/0.39}          & \multicolumn{2}{c}{2.07/0.39}          \\
maca.2   & \multicolumn{2}{c}{\textbf{2.73/0.49}} & \multicolumn{2}{c}{2.42/0.47} & \multicolumn{2}{c}{2.40/0.48} & \multicolumn{2}{c|}{2.59/0.38}          & \multicolumn{2}{c}{2.55/0.49}          \\
caps.    & \multicolumn{2}{c}{1.69/0.61}          & \multicolumn{2}{c}{1.55/0.60} & \multicolumn{2}{c}{1.52/0.60} & \multicolumn{2}{c|}{1.69/0.48}          & \multicolumn{2}{c}{\textbf{1.73/0.60}} \\
cand.      & \multicolumn{2}{c}{\textbf{2.61/0.24}} & \multicolumn{2}{c}{2.49/0.22} & \multicolumn{2}{c}{2.60/0.23} & \multicolumn{2}{c|}{1.80/0.12}          & \multicolumn{2}{c}{2.55/0.24}          \\
cash.      & \multicolumn{2}{c}{2.28/0.42}          & \multicolumn{2}{c}{2.04/0.40} & \multicolumn{2}{c}{2.01/0.39} & \multicolumn{2}{c|}{1.91/0.45}          & \multicolumn{2}{c}{\textbf{2.41/0.44}} \\
chew.  & \multicolumn{2}{c}{2.30/0.51}          & \multicolumn{2}{c}{2.17/0.47} & \multicolumn{2}{c}{2.12/0.46} & \multicolumn{2}{c|}{2.48/0.49}          & \multicolumn{2}{c}{\textbf{2.68/0.48}} \\
fryum       & \multicolumn{2}{c}{1.96/0.33}          & \multicolumn{2}{c}{1.91/0.30} & \multicolumn{2}{c}{1.88/0.29} & \multicolumn{2}{c|}{1.95/0.23}          & \multicolumn{2}{c}{\textbf{2.08/0.29}} \\
pipe. & \multicolumn{2}{c}{2.51/0.26}          & \multicolumn{2}{c}{2.23/0.34} & \multicolumn{2}{c}{2.31/0.34} & \multicolumn{2}{c|}{\textbf{3.04/0.36}} & \multicolumn{2}{c}{2.94/0.36}          \\ \midrule
mean     & \multicolumn{2}{c}{2.10/0.39}          & \multicolumn{2}{c}{1.85/0.37} & \multicolumn{2}{c}{1.86/0.37} & \multicolumn{2}{c|}{1.94/0.33}          & \multicolumn{2}{c}{\textbf{2.16/0.39}} \\ \bottomrule
\end{tabular}}
\caption{\textbf{Comparison of IS/IL on VisA dataset.} Best results are in \textbf{bold}.
}
\label{table:anomaly_generation_results}
\end{table}

%% file: figures/fig_tsne.tex
\begin{figure}[h!]
\centering\includegraphics[width=\linewidth]{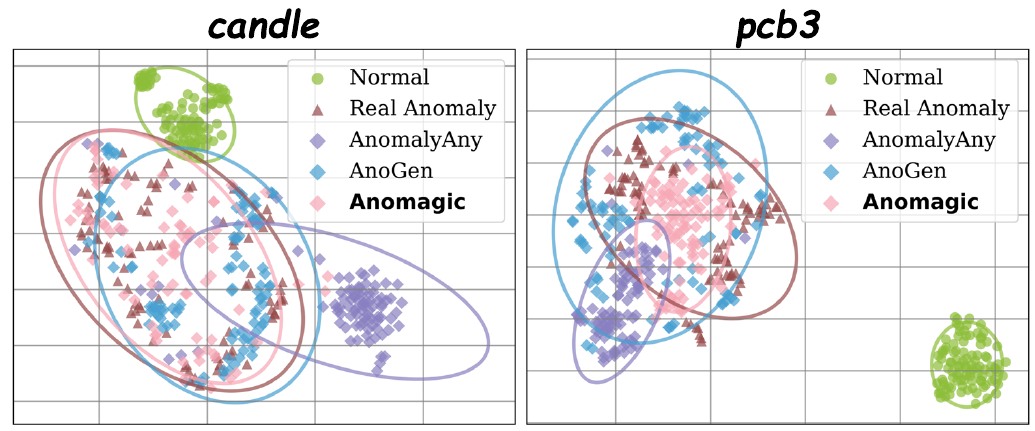}
\caption{\textbf{t-SNE visualization of marginal group distributions for ``candle'' and ``pcb3'' objects from  VisA.} 
} 
\label{fig:tsne}
\end{figure}

%% file: tables/table2.tex
\begin{table}[htbp]
\centering
\renewcommand{\arraystretch}{0.8}
\setlength{\tabcolsep}{10pt}
\begin{tabular}{ccccc}
\toprule
    {Method}                 & I-ROC       & I-F1           & PRO        & P-F1           \\ 
    \midrule
    {PatchCore}              & 95.10          & 94.10          & 91.20          & 44.70          \\
    {RD4AD}                  & 96.00          & 94.30          & 70.90          & 42.60          \\
    {Dinomaly}               & 98.90          & 96.20          & 95.30          & 48.60          \\ 
\midrule
{AnoGen}   &{\textbf{99.09}} &{96.55} &{95.62} &{52.61}     \\
                  DRAEM        & 99.03     & 96.58        & 95.59     & 51.94     \\
                  RealNet      & 99.03     & 96.75        & 95.70     & 52.87     \\
                  AnoAny   & 99.01     & 96.48        & 95.57     & 50.76     \\
                  Anomagic     & 99.08 & \textbf{96.77} & \textbf{95.92} & \textbf{54.00} \\
\bottomrule
\end{tabular}%
\caption{\textbf{Comparison of anomaly detection performance on VisA}. Below presents INP-Former++ augmented with selected anomaly generation methods. Best results are in \textbf{bold}.}
\label{tab:method_comparison}
\end{table}


%% file: tables/table3.tex
\begin{table}[ht]
    \centering
    \renewcommand{\arraystretch}{0.8}
    \setlength{\tabcolsep}{3.8pt}  
    \begin{tabular}{cccccccc}
        \toprule
        \multicolumn{2}{c}{Module} & \multicolumn{6}{c}{Metric} \\
        \cmidrule(r){1-2} \cmidrule(l){3-8}
        CPE & LoRA & IS & IL & I-ROC & I-F1 & PRO & P-F1 \\
        \midrule
         &  & 1.85 & 0.375 & {99.03} & 96.58 & 95.59 & 51.94 \\
        $\checkmark$ &  & \textbf{2.16} & \textbf{0.394} & 99.04 & 96.71 & 95.88 & 53.87 \\
        $\checkmark$ & $\checkmark$ & \textbf{2.16} & \textbf{0.394} & \textbf{99.07} & \textbf{96.77} & \textbf{95.92} & \textbf{54.00} \\
        \bottomrule
    \end{tabular}
        \caption{\textbf{Ablation Study on CPE and LoRA.} The method excludes both CPE and LoRA modules corresponds to DRAEM. \textbf{Bold} values indicate the best performance.} 
    \label{table:ablation_cpe_lora}
\end{table}

%% file: figures/fig_crossmodal_prompt.tex
\begin{figure}[th]
\centering\includegraphics[width=0.9\linewidth]{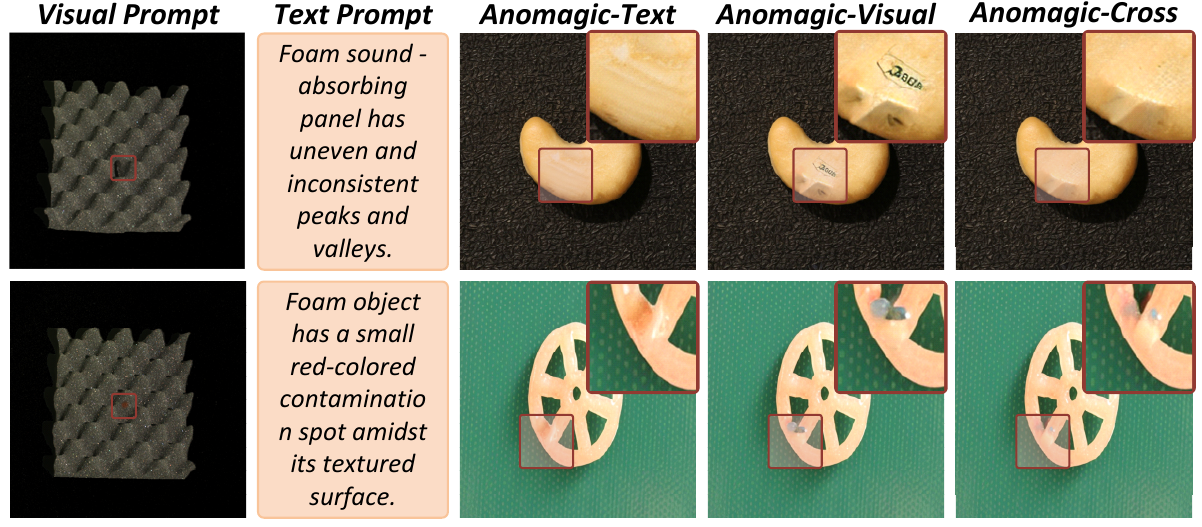}
\caption{\textbf{Visualization of anomaly generation results under unimodal and crossmodal prompts.} Our method effectively synthesizes realistic anomalies in both settings, with \ourmethod{}-Cross producing notably superior results.}
\label{fig:crossmodal_prompts}
\end{figure}

%% file: figures/fig_one_shot.tex
\begin{figure}[th]
\centering\includegraphics[width=0.9\linewidth]{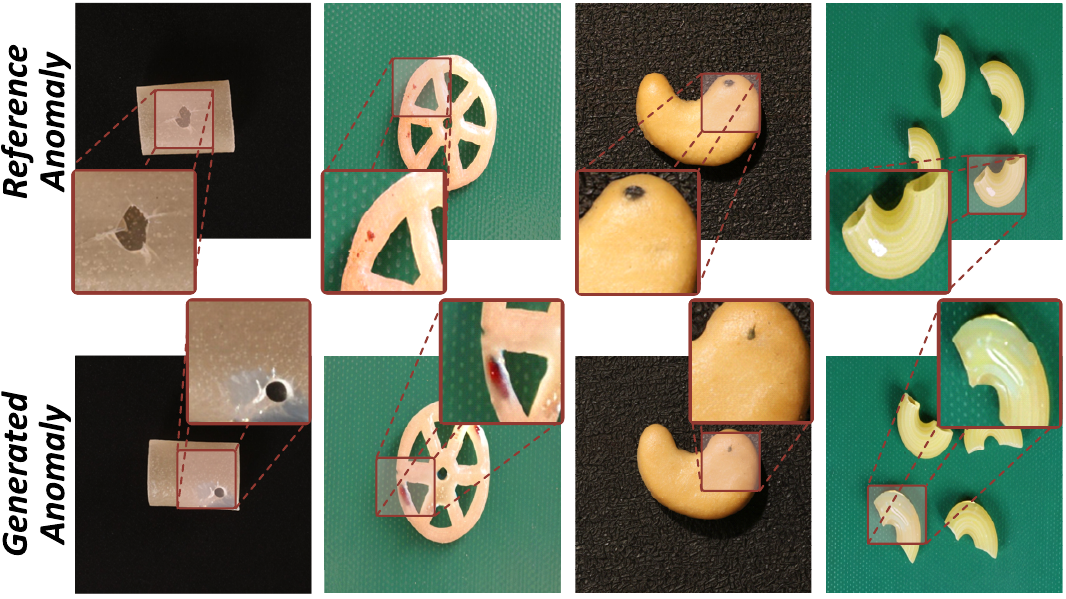}
\caption{\textbf{Visualization of anomaly generation results on VisA using unseen prompts from the same category.} Noting that \ourmethod{} is untrained on VisA, it can still generate realistic anomalies aligned with the provided prompts.}
\label{fig:one_shot}
\end{figure}

%% file: source/5-conclusion.tex
\section{Conclusion}
We have presented \ourmethod{}, a crossmodal prompt-driven, zero-shot anomaly synthesis method, together with \ourdataset{}, a dataset containing 12,987 anomaly–mask–caption triplets. Our prompt-guided inpainting mechanism produces high-fidelity, diverse anomalies that are precisely aligned with their masks. When integrated into INP‐Former++, these synthetic anomalies lead to SOTA detection performance. We further demonstrate \ourmethod{}’s flexibility in generating anomalies across a wide range of categories using both unimodal and crossmodal prompts, establishing it as a foundational tool for anomaly generation.

\noindent\textbf{Future Work:} We plan to augment \ourmethod{} and \ourdataset{} with fine‐grained control over anomaly attributes, such as geometric and material properties, to enable more customizable and targeted anomaly synthesis. To further validate the contrastive anomaly mask refinement strategy, we will compare boundary accuracy between refined and ground-truth masks, while assessing its robustness across a wider range of anomalies, including low-contrast cases.

%% file: source/Acknowledgments.tex
\section{Acknowledgments}

This work was supported by Fundamental Research Funds for the Central Universities (HUST: 2021GCRC058) and was part by the HPC Platform of Huazhong University of Science and Technology where the computation is completed.